\pdfoutput=1

\documentclass[11pt]{article}

\usepackage{acl}

\usepackage{times}
\usepackage{latexsym}

\usepackage[T1]{fontenc}

\usepackage[utf8]{inputenc}

\usepackage{microtype}

\usepackage{inconsolata}

\usepackage{hyperref}       
\usepackage{url}            
\usepackage{booktabs}       
\usepackage{amsfonts}       
\usepackage{nicefrac}       
\usepackage{microtype}      
\usepackage{xcolor}         

\usepackage{multirow}
\usepackage{paralist}
\usepackage{subcaption}
\usepackage{svg}
\usepackage{graphicx}
\usepackage{adjustbox}
\usepackage{amsmath}
\usepackage{amsthm}
\usepackage[capitalize]{cleveref}

\theoremstyle{definition}
\newtheorem{definition}{Definition}[section]

\let\P\relax
\DeclareMathOperator{\P}{\mathbb{P}}

\DeclareMathOperator{\similarity}{sim}

\newcommand{\calB}{\mathcal{B}}
\newcommand{\calD}{\mathcal{D}}

\newcommand{\calL}{\mathcal{L}}
\newcommand{\calM}{\mathcal{M}}

\newcommand{\calS}{\mathcal{S}}

\title{Synthetic Query Generation for Privacy-Preserving Deep Retrieval Systems using Differentially Private Language Models}

\author{Aldo Gael Carranza \\
  Stanford University \\
  \texttt{aldogael@stanford.edu} \\ \And
  Rezsa Farahani \\
  Google Inc. \\
  \texttt{farahani@google.com} \\ \And
  Natalia Ponomareva \\
  Google Research \\
  \texttt{nponomareva@google.com} \\ \AND
  Alex Kurakin \\
  Google DeepMind \\
  \texttt{kurakin@google.com} \\ \And
  Matthew Jagielski \\
  Google DeepMind \\
  \texttt{jagielski@google.com} \\ \And
  Milad Nasr \\
  Google DeepMind \\
  \texttt{srxzr@google.com} \\ }

\begin{document}
\maketitle
\begin{abstract}
We address the challenge of ensuring differential privacy (DP) guarantees in training deep retrieval systems. Training these systems often involves the use of contrastive-style losses, which are typically non-per-example decomposable, making them difficult to directly DP-train with since common techniques require per-example gradients. To address this issue, we propose an approach that prioritizes ensuring query privacy \textit{prior} to training a deep retrieval system. Our method employs DP language models (LMs) to generate private synthetic queries representative of the original data. These synthetic queries can be used in downstream retrieval system training without compromising privacy. Our approach demonstrates a significant enhancement in retrieval quality compared to direct DP-training, all while maintaining query-level privacy guarantees. This work highlights the potential of harnessing LMs to overcome limitations in standard DP-training methods.
\end{abstract}

\section{Introduction}

Deep retrieval systems have been widely adopted in many online services, from search to advertising, to match user queries to relevant recommendations
\citep{covington2016recsys, huang2020embedding}.
In many applications, candidate items for retrieval are often publicly available non-personal information in the sense that they do not contain any specific information related to any single user (e.g., articles, products, movies, ads). However, the input queries to retrieval systems can often contain user personal information.
Therefore, training deep retrieval systems on user data may enhance user experience through timely relevance, but it may also unintentionally compromise user privacy since neural network models have been demonstrated to implicitly memorize and leak sensitive user information in the training data \citep{carlini2019secret}. This raises privacy sensitivities around each stage of data collection, training, inference, and hosting these systems.
In this work, we seek to address the problem of ensuring
user query privacy guarantees in deep retrieval systems without significantly hindering their utility.

The standard approach to ensure the privacy of training data in many large-scale machine learning models is to directly introduce \textit{differential privacy} (DP) \citep{dwork2014algorithmic} guarantees during training \citep{abadi2016deep,ponomareva2023dp}.
These DP-training strategies provide guarantees by limiting the impact each individual data instance has on the overall model.
However, some models contain design elements that inherently hinder the ability to limit per-example contributions, and thus are more difficult to directly DP-train.
These include models with components that calculate batch statistics such as batch normalization layers \citep{ponomareva2023dp} and models with losses that \textit{cannot} be decomposed into per-example losses such as pairwise and contrastive-style losses \citep{huai2020pairwise, xue2021differentially}.

This limits the application of DP-training on deep retrieval systems since these systems typically use \textit{non-per-example decomposable contrastive-style losses} to train semantic neural representations of user queries and candidate items in order to facilitate efficient vector-based retrieval strategies.
The injected noise needed to achieve DP guarantees for these losses can scale with the number of candidate items that appear in the example-level loss computations, which can result in excessive retrieval quality degradation.
Hence, additional considerations are often needed to adapt DP-training to deep retrieval models for achieving an adequate privacy-performance tradeoff.

\begin{figure*}[ht]
  \centering
  \adjustbox{trim={.1\width} {.35\height} {.1\width} {.3\height},clip}{
    \includegraphics[width=\linewidth]{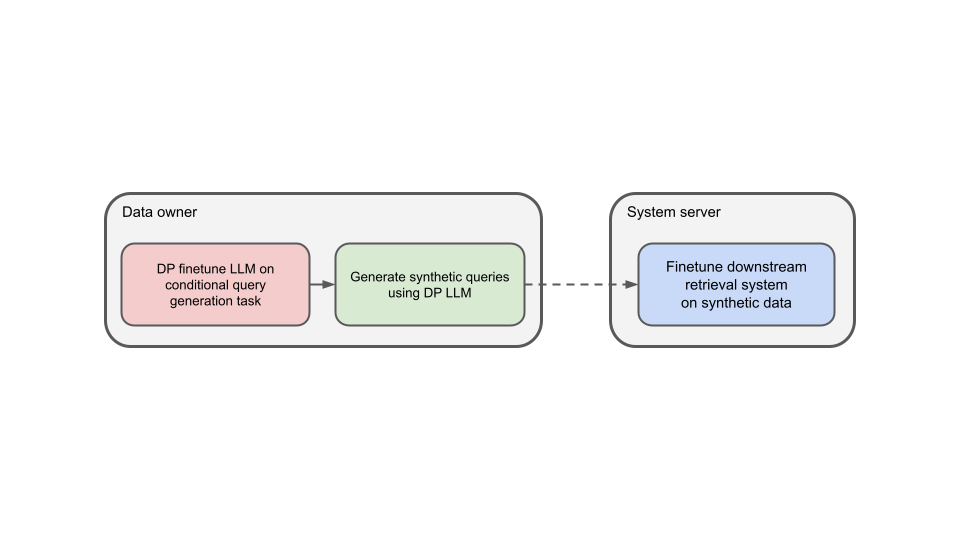}
  }
  \caption{Illustration of approach.}
  \label{fig:diagram}
\end{figure*}

In this work, we take an approach that ensures user query privacy \textit{prior} to training a deep retrieval system in order to circumvent the various issues with directly DP-training deep retrieval systems with a non-per-example decomposable contrastive-style loss. We build on the framework of \textit{synthetic data generation} using \textit{DP language models} (LMs) \citep{yue2022synthetic, mattern2022differentially} to develop an approach for private query sharing to train any downstream deep retrieval system with \textit{query-level privacy} guarantees with respect to the original training data. We empirically demonstrate considerable improvements in our approach on retrieval quality without compromising privacy guarantees compared to direct DP-training methods.
More generally, our work presents an initial study into a nascent opportunity to leverage exciting breakthroughs in LMs
to overcome crucial limitations in directly DP-training machine learning systems.

\section{Related Work}

\paragraph{Synthetic Generation using DP LMs}

Several recent studies \citep{yue2022synthetic, mattern2022differentially, mireshghallah2022privacy, putta2022differentially} have investigated the utility of private synthetic data from DP finetuned LMs in downstream tasks with pointwise losses, including text classification and semantic parsing. These works find that downstream models trained with private synthetic data outperform directly DP-trained models under the same privacy budget,
and non-private synthetic data generation can even improve performance
under no privacy constraints.
The reason is that DP synthetic data benefits from the injection of additional public information from the pretrained LMs.
Our work contributes another exploration of the advantages of private synthetic data generation for downstream training under a different learning paradigm with a non-per-example decomposable loss. In particular, our motivation is to achieve query privacy DP guarantees in deep retrieval systems with high utility.

\paragraph{DP-training under Non-per-example Decomposable Losses}
Figuring out better ways of DP-training models with non-per-example decomposable losses remains an active research topic. Research in this area has entirely focused on pairwise losses \citep{huai2020pairwise, xue2021differentially, kang2021towards}, introducing specialized algorithms under particular conditions like convexity, smoothness, and Lipshitz continuity to maintain a reasonable bound on sensitivity.
Our work presents a general-purpose approach without such additional assumptions for achieving some level of privacy for a system trained with a non-per-example decomposable loss.

\section{Background}

\subsection{Deep Retrieval}\label{sec:deep_retrieval}

\begin{figure}[ht]
  \centering
  \adjustbox{trim={.22\width} {.2\height} {.22\width} {.2\height},clip}{
    \includegraphics[width=1.6\linewidth]{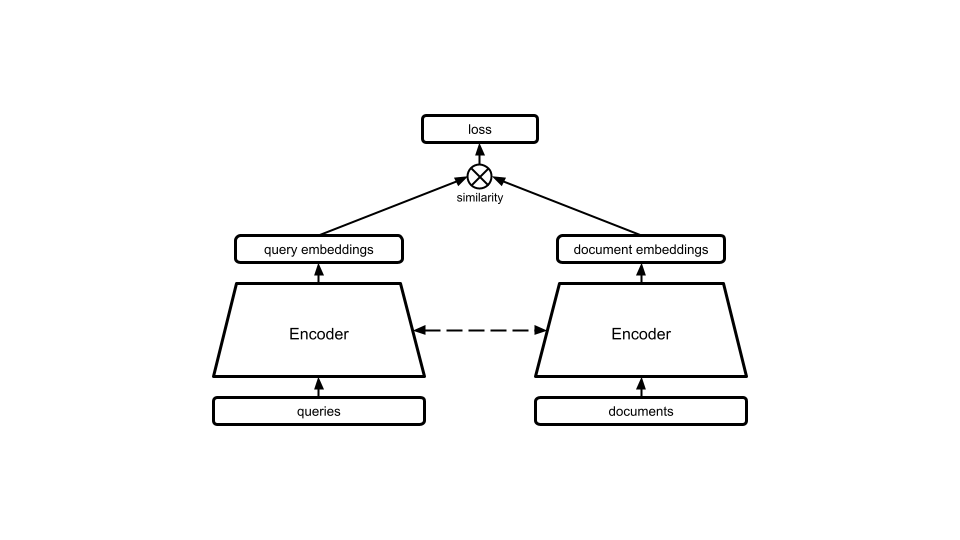}
  }
  \caption{Illustration of deep retrieval dual encoder model. Dashed lines connecting the query and document encoders represent the possibility of sharing the same encoder.}
  \label{fig:dual_encoder}
\end{figure}

Deep retrieval systems
have emerged as highly effective and scalable
information retrieval systems
to find candidate items
based on their semantic relevance to a specific query
\citep{huang2020embedding, ni2021large}.
These systems
typically consist of two neural encoders capable of generating rich, dense representations of queries and items (see Figure \ref{fig:dual_encoder}), which enable efficient \textit{approximate nearest neighbor search} methods \citep{guo2020accelerating}
to retrieve semantically relevant items to a given query.
Deep retrieval systems are typically trained on \textit{contrastive-style losses} that make use of two types of data examples: positive examples and negative examples. The positive examples help train the encoders into pulling relevant query-item pair embeddings close together in the embedding space, while negative examples help in
preventing embedding space collapse.
A popular choice for the loss function in deep retrieval is the \textit{in-batch softmax loss}, which makes memory-efficient use of items already loaded in a mini-batch as randomly sampled soft negatives \citep{gillick2019learning,karpukhin2020dense,qu2020rocketqa}.
In particular, given a training batch of query-item pairs $\{(q_i,d_i)\}_{i\in\calB}$, each $d_i$ is the positive item document for query $q_i$, and all other item documents $\{d_j\}_{j\neq i}$ within the batch are treated as the negatives. The in-batch softmax loss for each sample in the batch is
\begin{equation}\label{eq:InBatchSoftmaxLoss}
    \calL_i=-\log\frac{e^{\similarity(q_i,d_i)}}{\sum_{j\in\calB}e^{\similarity(q_i,d_j)}},
\end{equation}
where $\similarity(q_i,d_j)$ is the cosine similarity between the embeddings of $q_i$ and $d_j$ for any $i,j\in\calB$.
The larger and more diverse the batch, the better it is for representation learning.

\subsubsection{Privacy Risks of Deep Retrieval}

The neural encoders in deep retrieval systems are high-capacity models known to implicitly memorize sensitive information present in the training data \citep{carlini2019secret}. Such sensitive data can subsequently be extracted from these trained models  \citep{carlini2021extracting,lehman2021does}. 
Moreover, retrieval-augmented text generation systems, which utilize deep retrieval systems to aid text generation, have been demonstrated to be more susceptible to leaking private information from their private datastore compared to the language models trained on the private data \citep{huang2023privacy, zeng2024good}.
This underscores the increase in privacy risks associated with deep retrieval systems.

\subsection{Conditional Text Generation}

Conditional text generation is the task of generating a sequence of text given a specific prompt \citep{keskar2019ctrl, schick2021few}. Pre-trained generative LMs such as GPT-3 and T5 have been shown to be highly effective at generating high-quality text conditioned on various prompt inputs \citep{raffel2020exploring, brown2020language}. 
Given a context $c$, the probability distribution of a text sequence $x=(x_1,\dots,x_n)$ is decomposed as
\begin{math}
    \smash[tb]{p(x|c)=\prod_{i=1}^n p(x_i|x_1,\dots,x_{i-1},c)}.
\end{math}
A neural network $p_\theta$ is trained to model the conditional distributions. The model can then be used to generate a new sample $\smash[tb]{\tilde x=(\tilde x_1,\dots, \tilde x_m)}$ conditioned on a given context $c$ by sequentially sampling $\smash[tb]{p_\theta(\cdot|c), p_\theta(\cdot|\tilde x_1,c),\dots,p_\theta(\cdot|\tilde x_1,\dots,\tilde x_{m-1}, c)}$.
In this work, we model the distribution of query texts given item documents as contexts with a publicly pre-trained LM.

\subsection{Differential Privacy}\label{sec:DP}

Differential privacy (DP) has become a gold standard for ensuring data anonymization \citep{dwork2014algorithmic}. In this work, we make use of the following relaxed notion of differential privacy known as $(\epsilon,\delta)$-differential privacy.

\begin{definition}[Differential Privacy]
    A randomized algorithm $\calM: \mathcal{D} \to \mathcal{S}$ is $(\epsilon,\delta)$-differentially private if for all $S\subset\calS$ and for any two neighboring datasets $D, D'\in\calD$ that differ exactly by a single data point, we have that
    \begin{math}
        \P[\calM(D)\in S]\le e^\epsilon\P[\calM(D')\in S] + \delta.
    \end{math}
\end{definition}

Note that for capturing query-level differential privacy in retrieval datasets under this definition, neighboring datasets are datasets that differ by exactly one query.
This definition captures a privacy guarantee based on the indistinguishability of the presence of a single data point in a dataset.
The $\epsilon$ and $\delta$ parameter control the strength of this privacy guarantee, where smaller values correspond to stronger guarantees. A useful property of DP that is crucial to our approach is its \textit{post-processing property}~\citep{dwork2014algorithmic} which states that for any deterministic or randomized function $f$ defined over the range of the mechanism $\calM$, if $\calM$ satisfies $(\epsilon,\delta)$-DP, so does the composition $f\circ\calM$. The post-processing property ensures that arbitrary computations on the output of a DP mechanism do not incur any additional privacy loss.

\subsubsection{Differentially Private Training}\label{sec:DP-Training}

In the context of machine learning, DP can be used to protect the privacy of data used to train a model, preventing an adversary from inferring the presence of specific training examples. 
By far the most practical method of introducing DP to non-convex ML models involves the modification of the training process to limit the impact that each individual data instance has on the overall model, also referred to as \textit{DP-training} \citep{ponomareva2023dp}. The most popular methods for DP-training are gradient noise injection methods like differentially private stochastic gradient descent (DP-SGD) \citep{abadi2016deep}.
DP-SGD works by clipping per-example gradients to have norm no greater than $C$ and adding isotropic Gaussian noise $\mathcal{N}(0,\sigma^2C^2\mathbf{I})$ to the clipped gradients before aggregating and applying the gradient update to model weights.
The noise multiplier $\sigma$ is set based on the privacy parameters $\epsilon, \delta$, and it can be determined using a privacy accountant \citep{abadi2016deep}.

Clipping is done to bound gradient sensitivity,
which captures how much a single example can influence the trained model.
The specific value of $C$ does not actually affect the $(\epsilon,\delta)$-DP guarantee, since a larger value of $C$ means more noise will be added to compensate.
However, the primary challenge in setting the clipping norm is in finding the right balance to maximize utility.
If the clipping norm is set too low, it may overly constrain the gradients during training.
If the clipping norm is set too high, sensitivity is less controlled, and too much noise is added to the gradients. Both cases hinder the model's ability to learn and worsen utility.

\subsubsection{Limitations of Directly Differentially Private Training Retrieval Systems}\label{sec:dp_train_original}

Our work was primarily motivated by the fact
that DP-SGD is \textit{not} immediately compatible with the in-batch softmax loss used to train dual encoders.
The primary reason is that per-example gradients of this loss depend on not just the example in consideration but also all other examples in the batch.
Therefore, a single example can influence multiple per-example gradient computations which immediately implies an increased sensitivity of the gradient that scales with the batch size.
In DP-training, higher sensitivity means that more noise needs to be added to the gradient updates during training to achieve the same level of privacy guarantees, which leads to worse utility.

Moreover, DP-SGD provides guarantees on example-level privacy, and every example in this case contains a query and item. However, in this work, we are interested in achieving query-level privacy which should be easier than protecting both queries and items. Standard DP-SGD is not able to guarantee these less strict levels of privacy.

Lastly, a systems-level issue of DP-training on the in-batch softmax loss is that in order to take advantage of vectorization and parallelization strategies for computing per-example gradients more quickly \citep{subramani2021enabling}, each query-item example in a batch must be duplicated to be contained in every de facto example in the batch, leading to a quadratic increase in memory requirements. Given fixed memory resources, this necessitates significantly smaller batch sizes, which has an additional deleterious effect beyond gradient clipping and noising since effective representation learning under the in-batch softmax loss highly depends on the amount and diversity of in-batch examples.
Our approach of training with private synthetic queries precludes the above limitations when training a downstream dual encoder deep retrieval model.

\section{Approach}

We describe our general-purpose approach to obtain DP synthetic data for training a downstream deep retrieval system while ensuring query-level privacy on the original training data.

\paragraph{1) DP-training LM on Conditional Query Generation Task}

First, we obtain a suitable publicly pre-trained LM that has not been pre-trained on the queries in the private training data.
We use DP-Adafactor
to DP fine-tune the chosen LM with a conditional query generation task.
DP-Adafactor is merely an Adafactor optimizer \citep{shazeer2018adafactor} that receives clipped and noised gradients as per the DP-SGD algorithm \citep{abadi2016deep}.
The conditional query generation task is the following: given a query-item document pair $(q,d)$ in the training data,
the LM is fine-tuned to generate the target text ``$q$'' given input text ``$d$''.
Note that for larger LMs with billions of parameters, it is possible to leverage more parameter-efficient finetuning techniques \citep{lester2021power, hu2021lora} in order to overcome the high cost of training such large models.
The effect of parameter-efficient DP-finetuning on the quality of synthetically generated retrieval data is a subject of further study.

\paragraph{2) Synthetic Query Generation using DP LM}

Then, the DP fine-tuned LM is capable of generating synthetic queries that are representative of the real queries and relevant to the items.
For each item document $d$, we generate a matching synthetic query $\tilde q$ by providing the input ``generate\_query: $d$'' to the model.
This method allows for generating multiple synthetic queries from each document.
A synthetic training dataset is then constructed to be the set of original documents matched with their corresponding synthetic queries.

\paragraph{3) Training Dual Encoder with DP Synthetic Data}

Lastly, the synthetic data can then be shared securely for any subsequent training tasks without taking on any additional DP losses on the original queries, as guaranteed by the post-processing property of DP (see Section \ref{sec:DP}).
In particular, we can train a dual encoder model with the in-batch softmax loss (see Equation \ref{eq:InBatchSoftmaxLoss}) on the synthetic training data using standard non-private training methods, while still guaranteeing DP protection of the original queries.

\section{Experimental Setup}

\subsection{Datasets}

We use publicly available datasets for information retrieval tasks. For finetuning and evaluation, we consider the MSMARCO dataset \citep{bajaj2016ms}, which consists of nearly 533,000 query-document pairs of search data sampled from Bing search logs covering a broad range of domains and concepts. Additionally, we consider datasets in the BEIR benchmark suite \citep{thakur2021beir}, which contains information retrieval datasets across a variety of domains, for zero-shot evaluation.

\subsection{Synthetic Data Generation}

\subsubsection{Implementation Details}

\paragraph{Model Training}

For synthetic data generation, we trained various T5 LMs \citep{raffel2020exploring} with different sizes  $\{\text{Small, Base, Large, XL}\}$ and privacy guarantees $\epsilon\in\{3, 8, 16, \infty\}$ to generate synthetic queries given corresponding input documents from the MSMARCO dataset.
The T5 Small, Base, Large, XL models have around 60 million, 220 million, 770 million, 3 billion parameters, respectively. All experiments were performed on a TPU v4 chip.

\vspace{-.25em}
\paragraph{Hyperparameters}

We trained each LM over 30 epochs with batch size 1024 and set the maximum token length to be 384 for input documents and 128 for target queries.
We used the DP-Adam optimizer with a learning rate of 0.001 and clip norm of 0.1.
Following \citep{li2021large}, we set the privacy parameter $\delta=1/2n$ where $n$ is the training dataset size.
For sampling, we used a nucleus sampling strategy \citep{holtzman2019curious} with $p=0.8$.

The hyperparameters above were chosen from a hyperparameter search to identify the optimal hyperparameters for the T5-Small model, DP-finetuned on the MSMARCO training dataset. The optimal criteria were the highest BLEU scores achieved on a validation dataset.
We found that learning rate of $0.001$, clipping norm $0.1$, batch size $1024$, and epochs $30$ mostly resulted in the best model. We used these hyperparameters in all other T5 models. See Table \ref{tab:hyperparameters} for the hyperparameter grid.

\begin{table}[ht]
\centering
\footnotesize
    \caption{\small Hyperparameter grid.}
    \label{tab:hyperparameters}
    \begin{tabular}{c|c}
    \toprule
        Hyperparameter &  Values \\
    \midrule\midrule
        Token Lengths & Input: 384, Target: 128 \\
        Learning Rate & $0.001\cdot 2^{-k}$ for $k\in\{0, 1, 2, 3\}$ \\
        Clipping Norm & $\{0.1, 0.25, 0.5, 1\}$ \\
        Batch Size & $\{128, 256, 512, 1024\}$ \\
        Epochs & $\{10, 20, 30\}$ \\
    \bottomrule
    \end{tabular}
\end{table}

\subsubsection{Data Synthesis}

We used each DP-finetuned T5 LM to generate synthetic queries given documents from the original training data.
These pairs of synthetic queries and original documents constitute a new synthetic dataset.
For qualitative comparison, we provide an example in Table \ref{tab:synthetic_examples2} of an original query-document pairs and the synthetic queries generated under various model configurations and privacy levels.

\begin{table*}[ht]
\centering
\footnotesize
    \caption{\small Synthetic query example.}
    \vspace{-.75em}
    \label{tab:synthetic_examples2}
    \begin{tabular}{p{0.08\linewidth} p{0.06\linewidth} | p{0.77\linewidth}}
    \toprule
         \multicolumn{2}{l|}{Source} &  Text \\
    \midrule\midrule
       \multicolumn{2}{l|}{Document} & The main cast of the show: Mickey Mouse, Minnie Mouse, Donald Duck, Daisy Duck, Goofy, and Pluto star in the series, which focuses on interacting with the viewer to stimulate problem solving. \\
       \multicolumn{2}{l|}{Original Query} & characters from the Mickey Mouse clubhouse show \\
    \midrule
        \multirow{4}*{T5-Base} & $\epsilon=\infty$ &    the Mickey Mouse show cast \\
        & $\epsilon=16$ &   what is the most important characters in the series \\
        & $\epsilon=8$ &   what is in this series? \\
        & $\epsilon=3$ &   what is in this code for dfr1 \\
    \midrule
        T5-XL & $\epsilon=3$ & issue with show by mickey mouse \\
        T5-Large & $\epsilon=3$ & what is the most animated characters on disney cartoon show \\
        T5-Small & $\epsilon=3$ &  what is isn't a character will do a story \\
    \bottomrule
    \end{tabular}
\end{table*}

\subsubsection{Pretraining and Training Data Overlap}\label{app:overlap}

We note the importance that the pretrained LMs used to generate the synthetic data were not markedly trained on the original query data we seek to make private.
Otherwise, the privacy guarantees would be undermined since the models would have already seen the data.
To address this matter in our experiments, we conducted an analysis to determine the extent of overlap of the MSMARCO dataset on the pre-training data of T5 models, the C4 common crawl dataset \citep{raffel2020exploring}. We conducted multiple runs of selecting random subsets of 10,000 query and text pairs to determine if there was an exact match in the C4 dataset.

Our analysis determined that while a significant percentage of MSMARCO documents (\textasciitilde22\%) were exactly found in C4 on average, a negligible percentage (<1.9\%) of MSMARCO queries were exactly found in C4 on average. Moreover, the queries that were found tended to be generic search terms which could be considered public knowledge. Since we are interested in query-level privacy, we consider this level of dataset overlap acceptable to give reasonable guarantees of privacy.
In Section \ref{sec:empirical_privacy}, we provide a more extensive study of the empirical privacy guarantees of our training procedure.

\subsection{Downstream Retrieval System}

\subsubsection{Implementation Details}

\paragraph{Model Training}
For each data source (i.e., original MSMARCO data and synthetic datasets for various $\epsilon$ and model sizes), we train a separate dual encoder model on the in-batch softmax loss.
We utilize a separate pre-trained T5-Base encoder for both the query and document encoders, sharing parameters between them.
Similar to data synthesis, we use this kind of encoder to ensure that it is not significantly pretrained on the original queries.
We emphasize that the encoders of the retrieval models are distinct from the T5 models used to generate synthetic data.

\vspace{-.25em}
\paragraph{Hyperparameters}
For the hyperparameters in dual encoder model training, we used learning rate $0.001$, batch size $32$, epochs $5$, the maximum token length $384$ for documents and $128$ for queries. For the directly DP finetuning experiments, we used a clipping norm of $0.1$.

\subsubsection{Baseline Approach}

A baseline comparison of our approach will be to compare against a deep retrieval system that is directly DP-trained on the original data.
For direct DP-training, we used the same hyperparameters as above, but given the memory constraints discussed in Section \ref{sec:dp_train_original}, the batch size for DP-training a dual encoder model had to be significantly decreased to 32.
We do not experiment with different downstream deep retrieval models since
our intent is to compare general-purpose methods for achieving DP guarantees in retrieval systems.

\section{Evaluation}
\subsection{Evaluation on Retrieval Tasks}

We evaluate the retrieval models on the MSMARCO test data set and various other BEIR retrieval data sets for zero-shot evaluation.
We evaluate on the normalized discounted cumulative gain score over the top 10 predictions (NDCG@10) which measures the relevance and ranking quality of items in a recommendation list, considering both quality and position.
We also report the recall score over the top 10 predictions (Recall@10), which measures the percentage of times the ground truth recommendation appears in the top 10 predictions.
We report the evaluation results of a single training run.

\subsubsection{Search \& Retrieval Procedure}

Evaluation of the dual encoder retrieval models requires a query-document nearest neighbor search implementation for the inference stage.
For our experiments, we used the Scalable Nearest Neighbors (ScaNN) library, an open-source library
that provides a fast and scalable approximate nearest neighbor search procedure \citep{guo2020accelerating}.
The procedure was executed using ScaNN's brute force scoring and the inner product distance settings.

\subsubsection{MSMARCO Evaluation}

Table \ref{tab:msmarco_eval} shows the evaluation results on the MSMARCO test set for deep retrieval models regularly trained on the synthetically generated data under varying generative model configurations and varying privacy levels.
For our benchmark comparison, in Table \ref{tab:msmarco_eval_og} we display the evaluation results on for the deep retrieval models directly DP-trained on the original data under varying privacy levels.
In the top rows of both tables, we also provide another baseline reference evaluation of a dual encoder model regularly finetuned on the original data without any DP guarantees (i.e., $\epsilon=\infty$).

We observe that the retrieval model trained on synthetic data with DP significantly outperform retrieval trained with DP on original data.
As discussed in Section~\ref{sec:dp_train_original}, there are a number of challenges associated with training DP models with contrastive-style losses.
Our naive approach of implementing DP-training on contrastive loss likely explains poor utility of DP-training on original data. 
Additionally, our DP synthetic data essentially introduces additional public knowledge into the process, since we utilize a publicly pretrained LM.

Moreover, we found that the retrieval model trained with non-DP synthetic data outperformed the retrieval model trained on the original data. This suggests that synthetic data generation indeed augments the original data and to some extent improves generalization, whether it be through introducing additional public information or data cleaning.
In fact, data augmentation via synthetic data generation using language models for deep retrieval is an area of research that has gained significant interest in recent years \citep{dai2022promptagator, bonifacio2022inpars}.
We also observe that performance increases with increasing model size. This is consistent with similar prior results that demonstrate DP-SGD on over-parameterized models can perform significantly better than previously thought \citep{de2022unlocking, li2021large}.
Overall, we show that training with synthetic data from DP LMs is viable for achieving DP guarantees and efficiency in retrieval models.

\begin{table}[ht]
\centering
\footnotesize
    \caption{\small Evaluation of retrieval models. Top: Trained on DP synthetic data with varying $\epsilon$ and fixed model size T5-Base. Bottom: Trained on DP synthetic data with varying model size and fixed $\epsilon=3$.}
    \label{tab:msmarco_eval}
    \begin{tabular}{c|c|c|c}
    \toprule
         Source &  $\epsilon$ &  NDCG@10 & Recall@10 \\
    \midrule\midrule
       Original &    $\infty$ &   0.2525 &    0.4098 \\
    \midrule
        T5-Base &    $\infty$ &   0.2590 &    0.4192 \\
        T5-Base &          16 &   0.2027 &    0.3342 \\
        T5-Base &           8 &   0.1912 &    0.3196 \\
        T5-Base &           3 &   0.1830 &    0.3108 \\
    \midrule
      T5-XL    &          3 &   0.1854 &   0.3098 \\
      T5-Large &          3 &   0.1833 &   0.3094 \\
      T5-Base  &          3 &   0.1830 &   0.3108 \\
      T5-Small &          3 &   0.1346 &   0.2272 \\
    \bottomrule
    \end{tabular}
\end{table}
    
\begin{table}[ht]
\centering
\footnotesize
    \caption{\small Evaluation of retrieval models DP-trained directly on original data with varying $\epsilon$.}
    \label{tab:msmarco_eval_og}
    \begin{tabular}{c|c|c|c}
    \toprule
         Source &  $\epsilon$ &  NDCG@10  & Recall@10 \\
    \midrule\midrule
       Original &    $\infty$ &   0.2525  &    0.4098 \\
    \midrule
       Original &          16 &   0.0523  &    0.0862 \\
       Original &           8 &   0.0386  &    0.0649 \\
       Original &           3 &   0.0234  &    0.0388 \\
    \bottomrule
    \end{tabular}
\end{table}

\subsubsection{Zero-shot Evaluation}

\begin{table*}[ht]
\centering
\footnotesize
    \caption{\small Zero-shot evaluation of retrieval models trained on DP synthetic data vs. directly DP-trained retrieval models with $\epsilon=16$. Top table: NDCG@10. Bottom table: Recall@10.}
    \label{tab:zero_eval}
    {\setlength{\tabcolsep}{3pt}
    \begin{tabular}{c|c|ccccccccccc}
    \toprule
    \multirow{2}{*}{Source} & \multirow{2}{*}{$\epsilon$} & \multicolumn{11}{c}{NDCG@10} \\
                            &                             &  arguana &  cqadup &  dbpedia &    fiqa &  hotpot &  nfcorpus &   quora &  scidocs &  scifact &   covid &  touche \\
    \midrule\midrule
                   Original &                    $\infty$ &   0.2653 &  0.2659 &   0.3905 &  0.2257 &  0.5232 &    0.2974 &  0.8126 &   0.1857 &   0.4527 &  0.4971 &  0.2764 \\
    \midrule
                   Original &                           16 &   0.2132 &  0.0990 &   0.1272 &  0.0870 &  0.1422 &    0.1331 &  0.6856 &   0.0792 &   0.2051 &  0.3133 &  0.1185 \\
                    T5-Base &                          16 &   0.2757 &  0.2474 &   0.3728 &  0.2140 &  0.5122 &    0.2971 &  0.7850 &   0.1750 &   0.4645 &  0.4351 &  0.2547 \\
    \bottomrule
    \multicolumn{1}{c}{}
    \end{tabular}}
    {\setlength{\tabcolsep}{3pt}
    \begin{tabular}{c|c|ccccccccccc}
    \toprule
    \multirow{2}{*}{Source} &  \multirow{2}{*}{$\epsilon$} & \multicolumn{11}{c}{Recall@10} \\
                            &                             &  arguana &  cqadup &  dbpedia &    fiqa &  hotpot &  nfcorpus &   quora &  scidocs &  scifact &   covid &  touche \\
    \midrule\midrule
                   Original &                    $\infty$ &   0.5569 &  0.3368 &  0.1479 &  0.2440 &  0.3706 &  0.1013 &  0.8989 &  0.1071 &  0.5801 &  0.0116 &  0.0530 \\
    \midrule
                   Original &                          16 &   0.4388 &  0.1325 &  0.0313 &  0.0906 &  0.1108 &  0.0365 &  0.7762 &  0.0503 &  0.2969 &  0.0063 &  0.0149 \\
                    T5-Base &                          16 &   0.5768 &  0.3165 &  0.1217 &  0.2261 &  0.3398 &  0.1110 &  0.8763 &  0.1066 &  0.5848 &  0.0101 &  0.0489 \\
    \bottomrule
    \end{tabular}}
\end{table*}

We also evaluate the zero-shot generalization capabilities of a retrieval model trained on synthetic data. We compare against a retrieval model trained on the original data with no DP (i.e., $\epsilon=\infty$) and with $\epsilon=16$. See Table \ref{tab:zero_eval} for the results.
Again, our results demonstrate significant advantage of DP synthetic data compared to DP-training on original data, nearly matching and in some cases outperforming the non-DP results. This suggests that the benefits of synthetic data generation can outweigh the utility degradation of DP-training with reasonable levels of privacy, at least in zero-shot generalization tasks.

\subsection{Similarity between Synthetic and Original Datasets}

We compute measures of similarity between the synthetic data generated by the DP-trained T5 models against the original data.
Since the synthetic data is one-to-one generated from the original data, we can compute BLEU scores to evaluate similarity \citep{post2018call}.
We also compute the MAUVE scores, shown to be more capable of comparing similarity of text distributions \citep{pillutla2021mauve}.
See Table \ref{tab:similarity_scores} for the scores.
We observe that the non-DP finetune model generates synthetic data that is as similar as one could expect under these metrics, and there is a significant drop with finite $\epsilon$, with increasing similarity with higher $\epsilon$ and increasing model size.
By comparing the similarity scores with the retrieval evaluation results, we observe that while larger models lead to drastic improvements in the synthetic data similarity, the downstream retrieval performance sees comparatively more modest gains with increasing model size.

\begin{table}[ht]
\centering
\footnotesize
    \caption{\small Similarity scores of generated synthetic data. Top: Varying $\epsilon$ and fixed model size. Bottom: Varying model size with fixed $\epsilon=3$.}
    \label{tab:similarity_scores}
    \begin{tabular}{c|c|cc}
    \toprule
       Model &  $\epsilon$ &    BLEU &   MAUVE \\
    \midrule\midrule
     T5-Base &    $\infty$ &  0.2939 &  0.9763 \\
     T5-Base &          16 &  0.0984 &  0.3715 \\
     T5-Base &           8 &  0.0940 &  0.3431 \\
     T5-Base &           3 &  0.0856 &  0.2974 \\
    \midrule
    T5-XL &        3 &  0.1021 & 0.7117 \\
    T5-Large &        3 &  0.1096 & 0.6359 \\
     T5-Base &        3 &  0.0940 & 0.2974 \\
    T5-Small &        3 &  0.0436 & 0.2296 \\
    \bottomrule
    \end{tabular}

\end{table}

\subsection{Empirical Privacy}\label{sec:empirical_privacy}

The provable privacy provided by DP decays significantly as $\epsilon$ grows, but prior work has shown that even these large values can provide strong protection against state of the art privacy attacks~\citep{carlini2019secret, carlini2022membership, ponomareva2023dp}. To verify our training technique still follows this tendency, we evaluate here the empirical privacy leakage of DP-trained language models, using the \emph{canary exposure} metric introduced in \cite{carlini2019secret}. This technique is frequently used to evaluate empirical privacy~\citep{zanella2020analyzing, ramaswamy2020training, jagielski2022measuring}. To perform this test, we construct examples with private information, referred to as canaries, and introduce a subset of them into the original training data, and measure how likely the model is to output the injected canaries. In general, canary generation is a domain-dependent decision, so we design canaries for our retrieval application using the three following types of query-document pairs:
(random query, random 10-digit string), (random query, corresponding document + random 10-digit string), (random query, random document + random 10-digit string).
The secret part of each canary is the random 10-digit string.

We train the language model on this modified dataset using different DP guarantees, generate synthetic datasets, and assess canary exposure.
We conduct this experiment multiple times with different canaries and DP guarantees, averaging the metrics and reporting the results in Table~\ref{tab:privacy_leakage}. As anticipated, training without DP leads to significant leakage. Canaries repeated 10 times are frequently extractable, and those repeated 100 times are always extractable. However, our approach with a large $\epsilon$ of 16 prevents the model from leaking secrets and significantly increases the rank. Recent techniques for converting attack success rates to lower bounds on the $\epsilon$ parameter~\citep{stock2022defending} allow us to interpret these ranks as a lower bound of roughly 0.015 on $\epsilon$. This large gap is consistent with prior findings on the empirical privacy of DP-SGD on language models \citep{carlini2019secret, ponomareva2023dp}.

\begin{table}[ht]
\centering
\footnotesize
    \caption{\small Privacy leakage.}
    \label{tab:privacy_leakage}
    \begin{tabular}{c|c|cc|cc}
    \toprule
    \multirow{2}{*}{Model} & \multirow{2}{*}{$\epsilon$} & \multicolumn{2}{c|}{Repetition = 10} & \multicolumn{2}{c}{Repetition = 100} \\
            &           &    Rank &  Leaked &    Rank &  Leaked \\ 
    \midrule\midrule
    T5-Base &  $\infty$ &   1/100 &    67\% &   1/100 &   100\% \\
    T5-Base &        16 &  43/100 &     0\% &  32/100 &     0\% \\
    \bottomrule
    \end{tabular}
\end{table}

\section{Conclusion}

Our work focused on ensuring DP guarantees in training deep retrieval systems.
We discussed the limitations of DP-training such systems with the often used in-batch softmax loss function, which is non-per-example decomposable.
We introduce an approach of using DP LMs to generate private synthetic queries for downstream deep retrieval training.
This approach ensures theoretical guarantees on query-level privacy prior to downstream training, thereby bypassing some of the limitations of DP-training.
Furthermore, we empirically demonstrate that our approach improves retrieval quality compared to direct DP-training, without compromising query privacy. Our work highlights the potential of LMs to overcome crucial limitations in DP-training ML systems.

\paragraph{Limitations}

There are a few limitations to our approach. Firstly, while we observed that larger LMs generate higher quality synthetic queries, it is worth noting that training such large models may too computationally expensive. Exploring more parameter-efficient finetuning methods tailored to DP-training could mitigate the computational burden associated with training such larger models.
Secondly, it is necessary for the publicly pretrained LM utilized for generating synthetic queries to not have been pretrained on the original queries we aim to privatize. This imposes a constraint on the choice of pretrained LMs suitable for generating private synthetic queries.
Next, as observed in the synthetic example in Table \ref{tab:synthetic_examples2}, the DP-finetuned LMs can sometimes generate incoherent queries, which can limit the relevance and interpretability of the data.
Lastly, it is essential to recognize that our approach exclusively ensures query-level privacy. For achieving more general example-level privacy, it may be necessary to resort to other approaches or more conventional DP-training methods.

\paragraph{Risks \& Ethical Considerations}

Data privacy is a crucial consideration in the responsible development of personalized machine learning systems. Our work directly offers potential solutions to address issues of data privacy in deep retrieval systems. However, it is important to acknowledge the above limitations on privacy guarantees of our approach to prevent undesired privacy risks.

\vspace{.25em}
\section*{Acknowledgements}
We thank Rishabh Bansal, Manoj Reddy, Andreas Terzis, Sergei Vassilvitskii, Abhradeep Guha Thakurta, Shuang Song, Arthur Asuncion, and Heather Yoon for the helpful discussions. We also thank Jianmo Ni for their assistance in setting up the retrieval training pipeline. Finally, we appreciate the support and encouragement of YouTube Ads leadership Shobha Diwakar, Marija Mikic, and Ashish Gupta throughout this work.

\bibliography{main}


\end{document}